\documentclass{article}

\usepackage[nonatbib,preprint]{nips_2018}

\usepackage[utf8]{inputenc} 
\usepackage[T1]{fontenc}    
\usepackage{hyperref}       
\usepackage{url}            
\usepackage{booktabs}       
\usepackage{amsfonts}       
\usepackage{nicefrac}       
\usepackage{microtype}      
\usepackage{graphicx}
\usepackage{subfig}
\usepackage[skip=0pt]{caption}
\usepackage{multirow}
\usepackage{color}
\usepackage{bm}
\usepackage{dsfont}
\usepackage{amsmath,amsthm}
\graphicspath{{./fig/}}

\newcommand{\X}{\mathcal{X}}
\newcommand{\Y}{\mathcal{Y}}

\newcommand{\Set}{\mathcal{S}}
\newcommand{\x}{\bm{x}}
\newcommand{\w}{\bm{w}}
\newcommand{\z}{\bm{z}}
\newcommand{\E}{\mathbb{E}}
\newcommand{\1}{\mathds{1}}

\newtheorem{theorem}{Theorem}[section]
\newtheorem{proposition}[theorem]{Proposition}
\newtheorem{remark}[theorem]{Remark}

\title{Federated Learning with Non-IID Data}

\author{
Yue Zhao* \\ \texttt{yzhao727@gmail.com}
\And Meng Li \\ \texttt{meng.li@arm.com}
\And Liangzhen Lai \\ \texttt{liangzhen.lai@arm.com}
\And Naveen Suda \\ \texttt{naveen.suda@arm.com}
\And Damon Civin \\ \texttt{damon.civin@arm.com}
\And Vikas Chandra \\ \texttt{vikas.chandra@arm.com}
}

\begin{document}

\maketitle

\begin{abstract}
Federated learning enables resource-constrained edge compute devices, such as
mobile phones and IoT devices, to learn a shared model for prediction, while
keeping the training data local. This decentralized approach to train models
provides privacy, security, regulatory and economic benefits.
In this work, we focus on the statistical challenge of federated learning when local
data is non-IID. We first show that the accuracy of federated learning 
reduces significantly, by up to \textasciitilde55\% for neural networks trained for
highly skewed non-IID data, where each client device trains only on a single class of
data. We further show that this accuracy reduction can be explained by the weight
divergence, which can be quantified by the earth mover's distance (EMD) between
the distribution over classes on each device and the population distribution. As a solution,
we propose a strategy to improve training on non-IID data by creating a small subset of data
which is globally shared between all the edge devices. 
Experiments show that accuracy can be increased
by \textasciitilde30\% for the CIFAR-10 dataset with only 5\% globally shared data.

\end{abstract}
%

\section{Introduction}
\label{introduction}

Mobile devices have become the primary computing resource for billions of 
users worldwide and billions more IoT devices are expected to come online 
over the next few years. These devices generate a tremendous amount of 
valuable data and machine learning models trained using these data have 
the potential to improve the intelligence of many applications. But 
enabling these features on mobile devices usually requires globally shared
data on a server in order to train a satisfactory model. This may be 
impossible or undesirable from a privacy, security, regulatory or economic 
point of view. Therefore approaches that keep data on the device and share 
the model have become increasingly attractive. 

There has been great progress recently on inference on the
device~\cite{zhang2017hello, lai2018cmsis}.
{Federated learning} \cite{FL_Arxiv2017_McMahan, konevcny2015federated, google2017fl} 
offers an approach to do training locally as well. 
McMahan \textit{et al.} \cite{FL_Arxiv2017_McMahan} introduced the 
$FederatedAveraging (FedAvg)$ algorithm and demonstrated the robustness 
of $FedAvg$ to train convolutional neural networks (CNNs) on benchmark 
image classification datasets (e.g. MNIST \cite{Lecun98gradient-basedlearning} 
and CIFAR-10 \cite{Krizhevsky09learningmultiple}), and LSTM on a language 
dataset \cite{shakespeare}. 

A lot of recent research has been done on addressing the communication
challenges of federated learning, i.e., how to reduce  the communication cost of transferring
huge matrices of weights of deep networks, and the unexpected dropout or the
synchronization latency due to network connectivity, power, and computing
constraints. Bonawitz \textit{et al.} \cite{bonawitz2017practical} developed
an efficient secure aggregation protocol for federated learning, allowing a
server to perform computation of high-dimensional data from mobile devices.
Kone{\v{c}}n{\`y} \textit{et al.} \cite{konevcny2016federated} proposed
{structured updates} and {sketched updates} to reduce
communication costs by two orders of magnitude. Lin \textit{et al.} \cite{lin2017deep}
proposed {Deep Gradient Compression (DGC)} to reduce the communication
bandwidth by two orders of magnitude to train high-quality models.

Other than the communication challenge, federated learning also faces 
the statistical challenge. Federated learning relies on {stochastic gradient
descent (SGD)}, which is widely used in training deep networks with good empirical
performances~\cite{duchi2011adaptive, tielemandivide, kinga2015method, bottou2010large,
rakhlin2012making, dean2012large}. The IID sampling of the training data is
important to ensure that the stochastic gradient is an unbiased estimate of the
full gradient~\cite{bottou2010large,  rakhlin2012making, ghadimi2013stochastic}.
In practice, it is unrealistic to assume that the local data on each edge device 
is always IID. 
To approach the non-IID challenge, Smith \textit{et al.} \cite{smith2017federated}
proposed a {multi-task learning (MTL)} framework and developed {MOCHA}
to address system challenges in MTL. But this approach differs significantly from the
previous work on federated learning. McMahan \textit{et al.}~\cite{FL_Arxiv2017_McMahan}
have demonstrated that $FedAvg$ can work with certain non-IID data. However,
as we will show in Section~\ref{problem}, the accuracy of convolutional neural networks trained with
$FedAvg$ algorithm can reduce significantly, up to 11\% for MNIST, 51\% for CIFAR-10 and 55\% for 
keyword spotting (KWS) datasets, with highly skewed non-IID data.

To address this statistical challenge of federated learning, we 
show in Section \ref{theory} that the accuracy reduction
can be attributed to the {weight divergence}, which quantifies
the difference of weights from two different training processes with the same weight initialization. We then
prove that the weight divergence in training is bounded by the {earth
mover's distance (EMD)} between the distribution over classes on each device
(or client) and the population distribution. This bound is affected by the
learning rate, synchronization steps and gradients. Finally, in
Section \ref{solution}, we propose a {data-sharing} strategy to improve $FedAvg$ with non-IID
data by distributing a small amount of globally shared data containing examples from
each class. This introduces a trade-off between accuracy and centralization.
Experiments show that accuracy can be increased by 30\% on CIFAR-10
if we are willing to centralize and distribute 5\% shared data.

\section{FedAvg on Non-IID data}
\label{problem}
In this section, we demonstrate the accuracy reduction with $FedAvg$ on non-IID data
by training representative neural networks on three datasets. 

\subsection{Experimental Setup}

In this work, we use convolutional neural networks (CNNs) trained on 
MNIST~\cite{Lecun98gradient-basedlearning}, 
CIFAR-10~\cite{Krizhevsky09learningmultiple} and 
Speech commands datasets~\cite{warden2018speech}.
MNIST and CIFAR-10 are datasets for image classification tasks with 
10 output classes.
Speech commands dataset consists of 35 words each of 1 sec duration. 
To make it consistent, we use a subset of the data with 10 key words as 
keyword spotting (KWS) dataset. 
For each audio clip, we extract 10 MFCC features for a 30ms frame
with a stride of 20ms to generate 50x10 features, which is used for the 
neural network training. For MNIST and CIFAR-10, we use same CNN architectures
as~\cite{FL_Arxiv2017_McMahan}, where as for KWS, we use CNN architecture 
from~\cite{zhang2017hello}.

The training sets are evenly partitioned into 10 clients.
For IID setting, each client is randomly assigned a uniform distribution over
10 classes.
For non-IID setting, the data is sorted by class and divided to create two 
extreme cases: (a) 1-class non-IID, where each client receives data partition 
from only a single class, and (b) 2-class non-IID, where the sorted data is
divided into 20 partitions and each client is randomly assigned 2 partitions
from 2 classes.



We use the same notations for $FedAvg$ algorithm as~\cite{FL_Arxiv2017_McMahan}: B, the batch size and E, 
the number of local epochs. The following parameters are used for $FedAvg$: 
for MNIST, B = 10 and 100, E = 1 and 5, $\eta = 0.01$ and decay rate = 0.995; 
for CIFAR-10, B = 10 and 100, E = 1 and 5, $\eta = 0.1$ and decay rate = 0.992; 
for KWS, B = 10 and 50, E = 1 and 5, $\eta = 0.05$ and decay rate = 0.992. 
The learning rates are optimized for each dataset and exponentially decays over 
communication rounds. For SGD, the learning rate and decay rate are identical 
but B is 10 times larger. This is because the global model from $FedAvg$ is 
averaged across 10 clients at each synchronization. $FedAvg$ with IID data 
should be compared to SGD with shuffling data and a batch size $K$ times 
larger, where $K$ is the number of clients included at each 
synchronization of $FedAvg$.

\begin{figure}[t]
	\centering
	\subfloat{\includegraphics[width=1.0\linewidth]{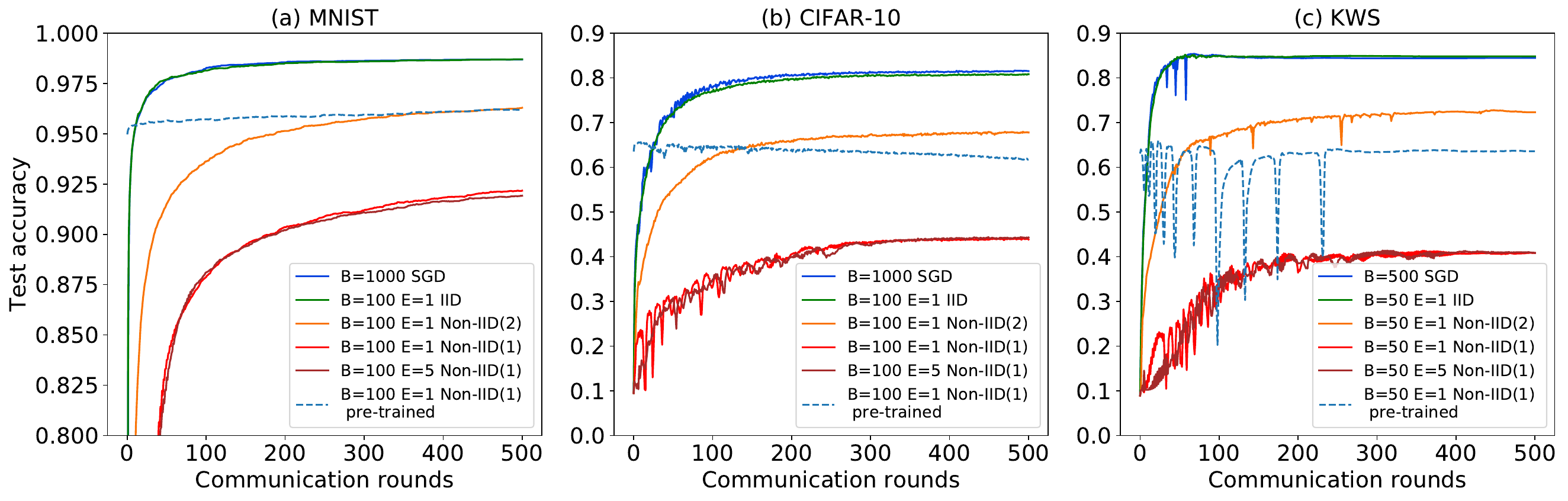}}\\
	\caption{Test accuracy over communication rounds of $FedAvg$ compared to SGD with IID and non-IID data of (a) MNIST (b) CIFAR-10 and (c) KWS datasets. Non-IID(2) represents the 2-class non-IID and non-IID(1) represents the 1-class non-IID. } 
	\label{problem:figure1}
\end{figure}

\subsection{Experimental Results}
For the IID experiments, the convergence curves of $FedAvg$ with a batch size of $B$ mostly overlap with the curves of SGD with $B \times 10$ for all the three datasets (Figure \ref{problem:figure1}). Only a small difference is observed for CIFAR-10 that $FedAvg$ with B = 10 converges to 82.62\% but SGD with B = 100 converges to 84.14\% (Figure \ref{apd:figure8}). Thus, $FedAvg$ achieves SGD-level test accuracy for IID data, which is consistent with the results in~\cite{FL_Arxiv2017_McMahan}. 
 
Significant reduction in the test accuracy is observed for $FedAvg$ on non-IID data compared to SGD with matched batch size (Figure \ref{problem:figure1} and \ref{apd:figure8}).
The accuracy reduction of non-IID data is summarized in Table \ref{problem:table1}. The maximum accuracy reduction occurs for the most extreme 1-class non-IID data. 
Moreover, a larger number of local epochs (E = 5) doesn't reduce the loss. The convergence curves mostly overlap for E=1 and E=5. Furthermore, the CNN models pre-trained by SGD doesn't learn from the $FedAvg$ training on non-IID data.
 For CIFAR-10, the accuracy drops when the pre-trained CNN is trained by $FedAvg$ on non-IID data. Thus, we demonstrate the reduction in the test accuracy of $FedAvg$ for non-IID data. 
 The test accuracy of all the experiments is summarized in Table \ref{apd:table3}. Note that the SGD accuracy reported in this paper are not state-of-the-art \cite{Lecun98gradient-basedlearning, krizhevsky2012imagenet, graham2014fractional, zhang2017hello}
 but the CNNs we train are sufficient for our goal to evaluate federated learning on non-IID data.

\begin{table}[h]
\centering
\caption{The reduction in the test accuracy of $FedAvg$ for non-IID data.}
\label{problem:table1}
\begin{tabular}{|c c c c c c|}
	\hline
	Non-IID & B & E & MNIST (\%)& CIFAR-10 (\%) & KWS (\%)\\
	\hline
	Non-IID(1) & large & 1 & 6.52 & 37.66 & 43.64\\
	\hline
	Non-IID(1) & large & 5 & 6.77 & 37.11 & 43.62\\
	\hline
	Non-IID(2) & large & 1 & 2.4 & 14.51 & 12.16\\
	\hline
	Non-IID(1) & small & 1 & 11.31 & 51.31 & 54.5\\
	\hline 
	Non-IID(2) & small & 1 & 1.77 & 15.61 & 15.07\\
	\hline
\end{tabular}
\end{table}

\section{Weight Divergence due to Non-IID Data}
\label{theory}

\begin{figure}[t]
	\centering
	\subfloat{\includegraphics[width=1.0\linewidth]{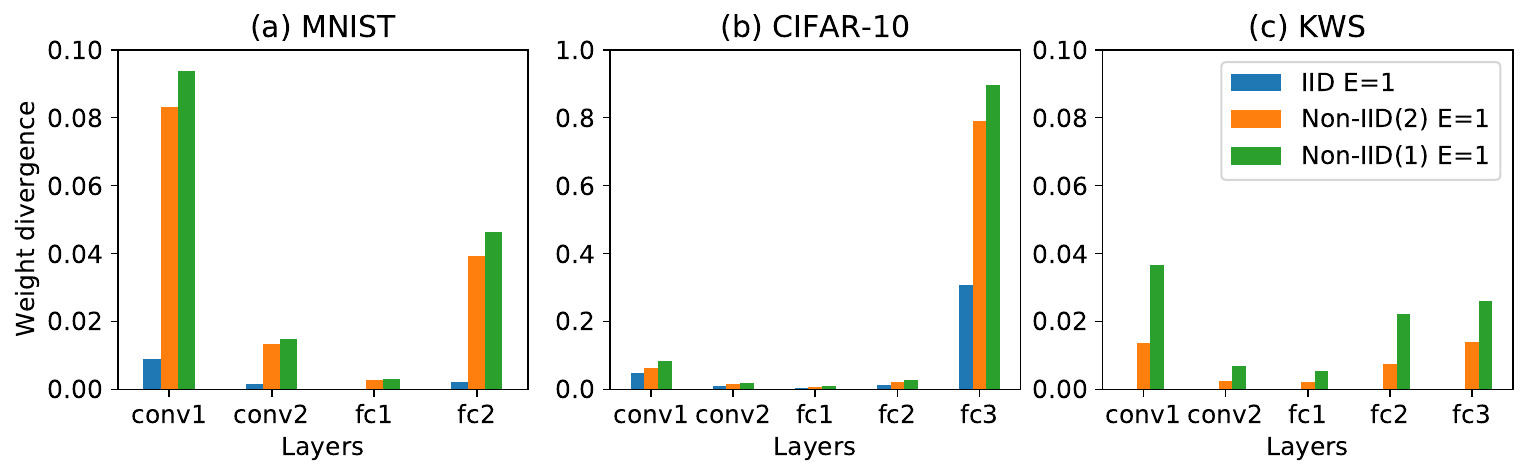}}
	\caption{Weight divergence of CNN layers for IID, 2-class non-IID and 1-class non-IID.}
	\label{problem:figure2}
\end{figure}

In Figure \ref{problem:figure1} and \ref{apd:figure8}, it is interesting to note that the reduction 
is less for the 2-class non-IID data than for the 1-class non-IID data. It 
indicates that the accuracy of $FedAvg$ may be affected by the exact data
distribution, i.e., the skewness of the data distribution.
Since the test accuracy is dictated by the trained weights,
another way to compare $FedAvg$ with SGD is to 
look at the difference of the weights relative to those of SGD, with 
the same weight initialization. It is termed as weight divergence and it 
can be computed by the following equation:
\begin{align}
\label{eq:df_weight_div}
weight\:divergence = || \w^{FedAvg} - \w^{SGD} || / || \w^{SGD} ||
\end{align}
As shown in Figure \ref{problem:figure2}, the weight divergence of all the layers increases as the data become more non-IID, from IID to 2-class non-IID to 1-class non-IID. Thus, an association between the weight divergence and the skewness of the data is expected. 
The accuracy reduction found in Section \ref{problem} can be understood in terms of {weight divergence}, which quantifies the difference of weights from two different training processes with the same weight initialization. In this section, we formally analyze the origin of the weight divergence. 
In Section \ref{theory:math}, we provide an illustrative example and a formal proposition to demonstrate that the root cause of the weight divergence is due to the distance between the data distribution on each client and the population distribution. Specifically, we find such distance can be evaluated with the {earth mover's distance} (EMD) between the distributions. 
Then, in Section \ref{theory:experiments}, we validate the proposition with experiments that demonstrate the impact of EMD on the weight divergence and test accuracy of $FedAvg$.

\subsection{Mathematical demonstration}
\label{theory:math}
We formally define the problem of federated learning and analyze the origin of the weight divergence. We consider a $C$ class classification problem defined over a compact space $\X$ and a label space $\Y = [C]$, where $[C] = \{1, \ldots, C\}$. The data point $\{\x, y\}$ distributes over $\X \times \Y$ following the distribution $p$. A function $f: \X \rightarrow \Set$ maps $\x$ to the probability simplex $\Set$, where $\Set = \{\z| \sum_{i=1}^{C} z_i = 1, z_i \geq 0, \forall i \in [C]\}$ with $f_i$ denoting the probability for the $i$th class. $f$ is parameterized over the hypothesis class $\w$, i.e., the weight of the neural network. We define the population loss $\ell(\w)$ with the widely used cross-entropy loss as
\begin{align*}
  \ell(\w) = \E_{\x, y \sim p}[\sum_{i=1}^{C} \1_{y=i} \log f_i(\x, \w)] = \sum_{i=1}^{C} p(y=i) \E_{\x|y=i}[\log f_i(\x, \w)].
\end{align*}
To simplify the analysis, we ignore the generalization error and assume the population loss is optimized directly. Therefore, the learning problem becomes
\begin{align*}
  \min_{\w} \sum_{i=1}^{C} p(y=i) \E_{\x|y=i}[\log f_i(\x, \w)].
\end{align*}
To determine $\w$, SGD solves the optimization problem iteratively. Let $\w^{(c)}_t$ denotes the weight after $t$-th update in the centralized setting. Then, centralized SGD performs the following update:
\begin{align*}
    \w^{(c)}_{t} = \w^{(c)}_{t-1} - \eta \nabla_{\w} \ell(\w^{(c)}_{t-1}) = \w^{(c)}_{t-1} - \eta \sum_{i=1}^{C} p(y=i) \nabla_{\w} \E_{\x|y=i}[\log f_i(\x, \w^{(c)}_{t-1})].
\end{align*}

In federated learning, we assume there are $K$ clients. Let $n^{(k)}$ denote the amount of data and $p^{(k)}$ denote the data distribution on client $k \in [K]$. On each client, local SGD is conducted separately. At iteration $t$ on client $k \in [K]$, local SGD performs
\begin{align*}
    \w^{(k)}_{t} = \w^{(k)}_{t-1} - \eta \sum_{i=1}^{C}  p^{(k)}(y=i) \nabla_{\w} \E_{\x|y=i}[\log f_i(\x, \w^{(k)}_{t-1})]. 
\end{align*}
Then, assume the synchronization is conducted every $T$ steps and let $\w^{(f)}_{mT}$ denote the weight calculated after the $m$-th synchronization, then, we have
\begin{align*}
    \w^{(f)}_{mT} = \sum_{k=1}^{K} \frac{n^{(k)}}{\sum_{k=1}^K n^{(k)}} \w^{(k)}_{mT}.
\end{align*}

\begin{figure}[!htb]
    \centering
    \includegraphics[width=\linewidth]{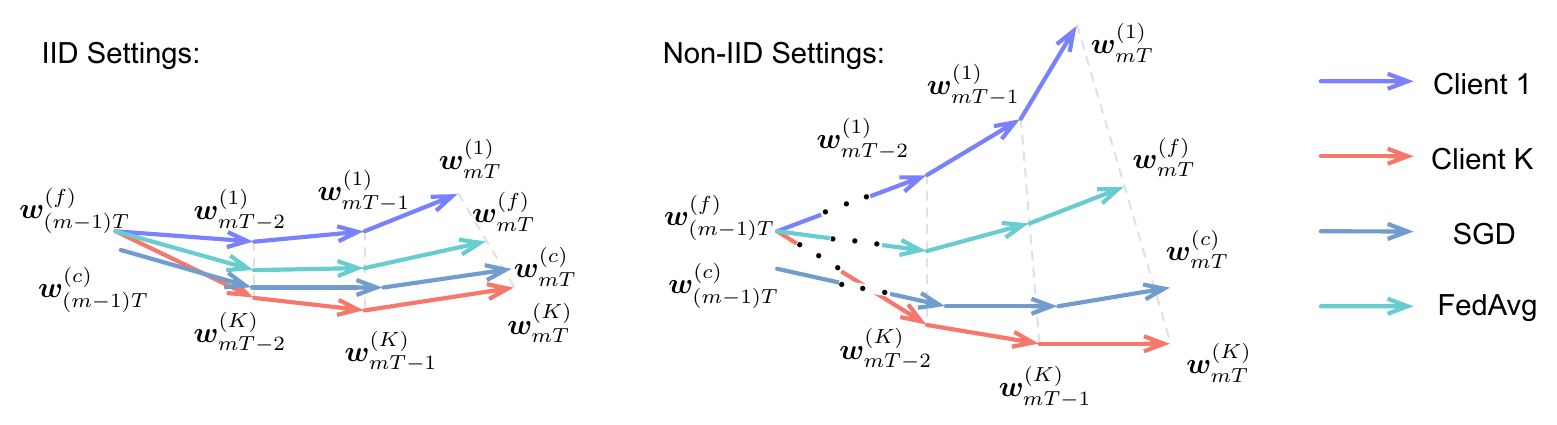}
    \caption{Illustration of the weight divergence for federated learning with IID and non-IID data.}
    \label{fig:weight_divergence}
\end{figure}

The divergence between $\w^{(f)}_{mT}$ and $\w^{(c)}_{mT}$ can be understood with the illustration in Figure~\ref{fig:weight_divergence}.
When data is IID, for each client $k$, the divergence between $\w^{(k)}_{t}$ and $\w^{(c)}_{t}$ is small and 
after the $m$-th synchronization, $\w^{(f)}_{mt}$ is still close to $\w^{(c)}_{mt}$.
When data is non-IID, for each client $k$, due to the distance between the data distribution,
the divergence between $\w^{(k)}_{t}$ and $\w^{(c)}_{t}$ becomes much larger and accumulates very fast,
which makes the divergence between $\w^{(f)}_{mT}$ and $\w^{(c)}_{mT}$ much larger.
To formally bound the weight divergence between $\w^{(f)}_{mT}$ and $\w^{(c)}_{mT}$, we have the following proposition.

\begin{proposition}
	\label{prop:weight_div}
    Given $K$ clients, each with $n^{(k)}$ i.i.d samples following distribution $p^{(k)}$ for client $k \in [K]$. If $\nabla_{\w} \E_{\x|y=i} \log f_i(\x, \w)$ is $\lambda_{\x|y=i}$-Lipschitz for each class $i \in [C]$ and the synchronization is conducted every $T$ steps, then, we have the following inequality for the weight divergence after the $m$-th synchronization,
	\begin{align}
	\label{eq:weight_div}
  		|| \w^{(f)}_{mT} - \w^{(c)}_{mT} || & \leq \sum_{k=1}^{K} \frac{n^{(k)}}{\sum_{k=1}^{K} n^{(k)}} (a^{(k)})^{T} || \w^{(f)}_{(m-1)T} - \w^{(c)}_{(m-1)T} || \notag\\
                                            & \qquad + \eta \sum_{k=1}^{K} \frac{n^{(k)}}{\sum_{k=1}^{K} n^{(k)}} \sum_{i=1}^{C} || p^{(k)}(y=i) - p(y=i) || \sum_{j=1}^{T-1} (a^{(k)})^{j} g_{max}(\w^{(c)}_{mT-1-k}),
	\end{align}
	where $g_{max}(\w) = \max_{i=1}^{C} || \nabla_{\w} \E_{\x|y=i} \log f_i(\x, \w) ||$ and $a^{(k)} = 1 + \eta \sum_{i=1}^{C} p^{(k)}(y=i) \lambda_{\x|y=i}$.
\end{proposition}

Detailed proof of Proposition~\ref{prop:weight_div} can be found in Appendix \ref{apd:deriv}. Based on Proposition~\ref{prop:weight_div}, we have the following remarks.

\begin{remark}
	\label{rmk:remark1}
	The weight divergence after the $m$-th synchronization mainly comes from two parts, including the weight divergence after the $(m-1)$-th divergence, i.e., $|| \w^{(f)}_{(m-1)T} - \w^{(c)}_{(m-1)T} ||$, and the weight divergence induced by the probability distance for the data distribution on client $k$ compared with the actual distribution for the whole population, i.e., $\sum_{i=1}^{C} || p^{(k)}(y=i) - p(y=i) ||$.
\end{remark}

\begin{remark}
	\label{rmk:remark2}
	The weight divergence after the $(m-1)$-th synchronization is amplified by $\sum_{k=1}^{K} \frac{n^{(k)} (a^{(k)})^{T}}{\sum_{k=1}^{K} n^{(k)}}$. Since $a^{(k)} \geq 1$, $\sum_{k=1}^{K} \frac{n^{(k)} (a^{(k)})^{T}}{\sum_{k=1}^{K} n^{(k)}} \geq 1$. Hence, if different clients start from different initial $\w$ in the federated learning, then, even if the data is IID, large weight divergence will still be encountered, which leads to degraded accuracy.
\end{remark}

\begin{remark}
	\label{rmk:remark3}
  	When all the clients start from the same initialization as the centralized settings, $\sum_{i=1}^{C} ||p^{(k)}(y=i) - p(y=i)||$ becomes the root cause of the weight divergence. This term is EMD between the data distribution on client $k$ and the population distribution, when the distance measurement is defined as $|| p^{(k)}(y=i) - p(y=i) ||$. The impact of EMD is affected by the learning rate $\eta$, the number of steps before synchronization $T$, and the gradient $g_{max}(\w^{(c)}_{mT-1-k})$. 
\end{remark}

Based on Proposition \ref{prop:weight_div}, we validate that EMD is a good metric to quantify
the weight divergence and thus the test accuracy of $FedAvg$ with non-IID data 
in Section~\ref{theory:experiments}.

\subsection{Experimental Validation}
\label{theory:experiments}
\subsubsection{Experimental Setup}
The training set is sorted and partitioned into 10 clients, $M$ examples per client. Eight values are chosen for EMD listed in Table \ref{theory:table1}. Because there may exist various distributions for one EMD, we aim to generate five distributions to compute the average and variations of the weight divergence and the test accuracy.
First, one probability distribution $P$ over 10 classes is generated for one EMD. Based on $M$ and $P$, we can compute the number of examples over 10 classes for one client. Second, a new distribution $P'$ is generated by shifting the 10 probabilities of $P$ by 1 element. The number of examples for the second client can be computed based on $P'$. This procedure is repeated for the other 8 clients. Thus, all the 10 clients have a distribution of M examples over 10 classes and each example is only used once. Finally, the above two steps are repeated 5 times to generate 5 distributions for each EMD. 
The CNNs are trained by $FedAvg$ over 500 communication rounds on the data processed from the above procedures. There are the key parameters used for training: for MNIST, B = 100, E = 1, $\eta$ = 0.01, decay rate = 0.995; for CIFAR-10, B = 100, E = 1, $\eta$ = 0.1, decay rate = 0.992; for KWS, B = 50, E = 1, $\eta$ = 0.05, decay rate = 0.992. The weight divergence is computed according to Eq. ~\eqref{eq:df_weight_div} after 1 synchronization (i.e., 1 communication round).
\subsubsection{Weight Divergence vs. EMD}
\begin{figure}[b]
  \centering
  \subfloat{\includegraphics[width=1.0\linewidth]{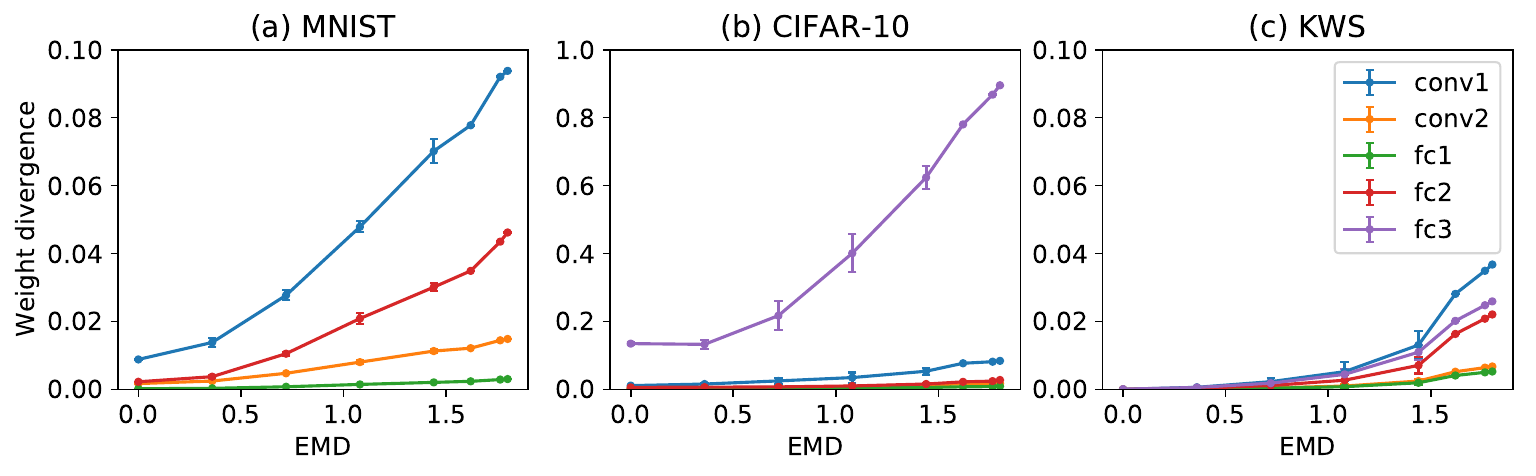}}
  \caption{Weight divergence vs. EMD across CNN layers on (a) MNIST, (b) CIFAR-10 and (c) KWS datasets. The mean value and standard deviation are computed over 5 distributions for each EMD.}
  \label{theory:figure3}
\end{figure}
The mean and standard deviation of the weight divergence are computed over 5 distributions for each EMD. For all the three datasets, the weight divergence of each layer increases with EMD as shown in Figure~\ref{theory:figure3}. The initial weights are identical for all SGD, IID and non-IID experiments on each dataset. Thus, according to Remark~\ref{rmk:remark1}, the weight divergence after 1 synchronization is not affected by the $(m-1)$-th divergence, $|| \w^{(f)}_{(m-1)T} - \w^{(c)}_{(m-1)T} ||$, because it is zero when $m = 1$. Therefore, the results in Figure \ref{theory:figure3} support Proposition~\ref{prop:weight_div} that the bound of weight divergence is affected by EMD. This effect is more significant in the first convolutional layer and the last fully connected layer. Moreover, the maximum weight divergence for CIFAR-10 is significantly higher than that for MNIST and KWS, which is affected by the gradient term in Eq.~\eqref{eq:weight_div} due to the problem itself and different CNN architectures. Note that the initial weights are also identical across the clients to avoid accuracy loss according to Remark~\ref{rmk:remark2}, which is consistent with the significant increase in the loss when averaging models with different initialization \cite{FL_Arxiv2017_McMahan, goodfellow2014qualitatively}.

\subsubsection{Test Accuracy vs. EMD}
\begin{figure}[h]
  \centering
  \subfloat{\includegraphics[width=.435\linewidth]{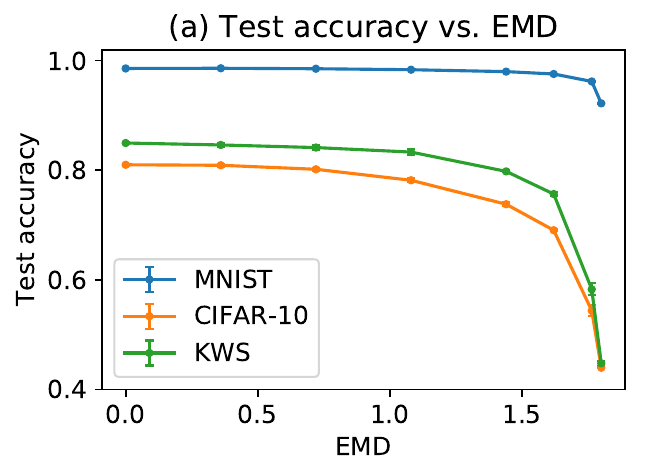}}
  \subfloat{\includegraphics[width=.5\linewidth]{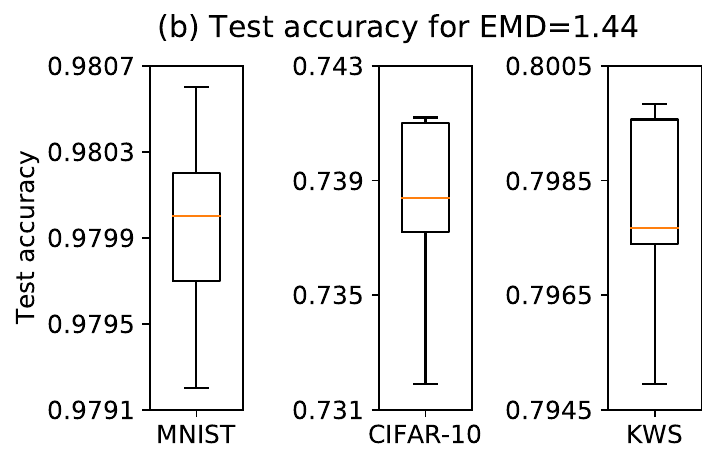}}
  \caption{(a) Test accuracy vs. EMD for $FedAvg$ and (b) boxplots of weight divergence when EMD = 1.44 for MNIST, CIFAR-10 and KWS datasets. The mean and standard deviation are computed over 5 distributions for each EMD.}
  \label{theory:figure4}
\end{figure}

The mean and standard deviation of the test accuracy are computed over the same 5 distributions for each EMD. The results are summarized in Table \ref{theory:table1} and are plotted against EMD in Figure \ref{theory:figure4}. For all the three datasets, the test accuracy decreases with EMD. The decreasing rate is relatively small at first and becomes larger as the data becomes more non-IID. So there is a trade-off between balancing non-IID data towards IID and improving the accuracy of $FedAvg$. The error bars on the plots represent the variations of test accuracy due to various distributions for each EMD. To take a closer look at the variations, the boxplots show the test accuracy across 5 runs when EMD = 1.44. Moreover, Table \ref{theory:table1} shows that the maximum variation is less than 0.086\% for MNIST, 2\% for CIFAR-10, and 1\%
for KWS, respectively. Thus, the accuracy is affected by EMD instead of the underlying distributions. It indicates that EMD can be used to estimate the accuracy of $FedAvg$ given a distribution of data. In addition, the maximum reduction in the accuracy is 6.53\% for MNIST, 37.03\% for CIFAR-10 and 40.21\% for KWS, respectively. This difference may be affected by the gradient in Eq.~\eqref{eq:weight_div} depending on the problem itself. 

\begin{table}[ht]
	\fontsize{8}{9}\selectfont
	\centering
	\caption{The mean and standard deviation of the test accuracy of $FedAvg$ over 5 distributions. The standard deviation is very small compared to the scale of the mean value.}.
	\label{theory:table1}
	\resizebox{\textwidth}{!}{%
		\begin{tabular}{|c |c | c c c c c c c c|}
			\hline
			\multicolumn{2}{|c|}{Earth mover's distance (EMD)} & 0 & 0.36 & 0.72 & 1.08 & 1.44 & 1.62 & 1.764 & 1.8\\
			\hline
			\multirow{2}{3em}{MNIST} & mean & 0.9857 & 0.9860 & 0.9852 & 0.9835 & 0.9799 & 0.9756 & 0.962 & 0.922\\
			& std ($\times 10^{-4}$) & 6.431 & 2.939 & 4.604 & 4.308 & 4.716 & 8.085 & 8.232 & 1.939\\
			\hline
			\multirow{2}{3em}{CIFAR-10} & mean & 0.8099 & 0.8090 & 0.8017 & 0.7817 & 0.7379 & 0.6905 & 0.5438 & 0.4396\\
			& std ($\times 10^{-3}$) & 2.06 & 2.694 & 2.645 & 3.622 & 3.383 & 2.048 & 9.655 & 1.068\\
			\hline
			\multirow{3}{3em}{KWS} & mean & 0.8496 & 0.8461 & 0.8413 & 0.8331 & 0.7979 & 0.7565 & 0.5827 & 0.4475\\
			& std ($\times 10^{-3}$) & 1.337 & 3.930 & 4.410 & 5.387 & 1.763 & 3.329 & 1.078 & 4.464\\
			\hline
	\end{tabular}}
\end{table}

\section{Proposed Solution}
\label{solution}
In this section, we propose a {data-sharing} strategy to improve $FedAvg$ with non-IID data by creating a small subset of data which is globally shared between all the edge devices. Experiments show that test accuracy can be increased by \textasciitilde30\% on CIFAR-10 dataset with only 5\% globally shared data.
\subsection{Motivation}
As shown in Figure \ref{theory:figure4}, the test accuracy falls sharply with respect to EMD beyond a certain threshold. Thus, for highly skewed non-IID data, we can significantly increase the test accuracy by slightly reducing EMD. As we have no control on the clients' data, we can distribute a small subset of global data containing a uniform distribution over classes from the cloud to the clients. This fits in with the initialization stage of a typical federated learning setting. In addition, instead of distributing a model with random weights, a warm-up model can be trained on the globally shared data and distributed to the clients. Because the globally shared data can reduce EMD for the clients, the test accuracy is expected to improve. 

\begin{figure}[h]
\begin{minipage}{0.38\textwidth}
	\centering
	\includegraphics[width=1.0\linewidth]{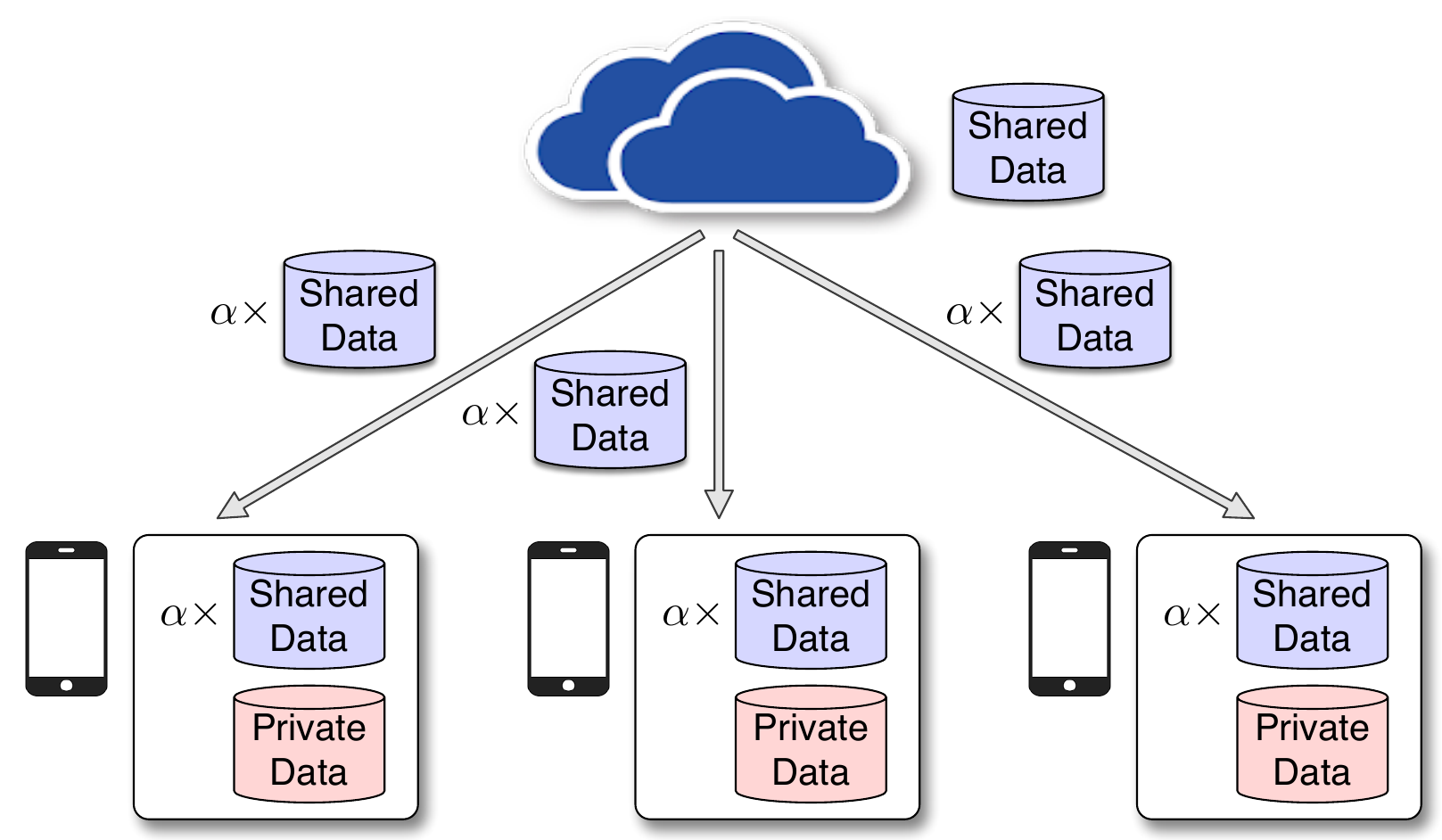}
	\caption{Illustration of the {data-sharing} strategy.}\label{sol:figure6}
\end{minipage}\hfill
\begin{minipage}{0.6\textwidth}
	\centering
	\includegraphics[width=1.0\linewidth]{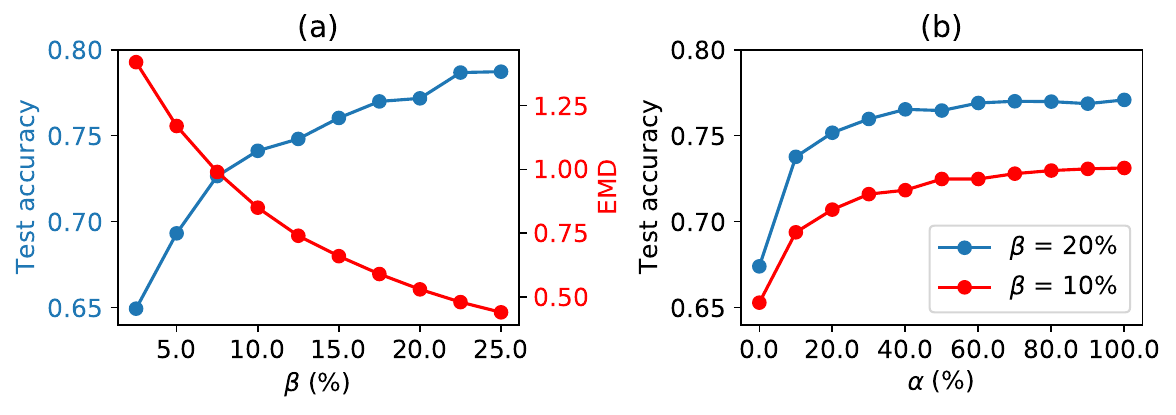}
	\caption{(a) Test accuracy and EMD vs. $\beta$ (b) Test accuracy vs. the distributed fraction $\alpha$}
	\label{sol:figure7}
\end{minipage}\hfill
\end{figure}

\subsection{Data-sharing Strategy}
Herein, we propose a {data-sharing} strategy in the federated learning setting as illustrated in Figure \ref{sol:figure6}. A {globally shared dataset} $G$ that consists of a uniform distribution over classes is centralized in the cloud. At the initialization stage of $FedAvg$, the warm-up model trained on $G$ and a random $\alpha$ portion of $G$ are distributed to each client. The local model of each client is trained on the shared data from $G$ together with the private data from each client. The cloud then aggregates the local models from the clients to train a global model with $FedAvg$. There are two trade-offs: (a) the trade-off between the test accuracy and the size of $G$, which is quantified as $\beta = \frac{||G||}{||D||} \times 100\%$, where $D$ represents the total data from the clients (b) the trade-off between the test accuracy and $\alpha$. The following experiments are performed on CIFAR-10 to address these two trade-offs.

The CIFAR-10 training set is partitioned into two parts, the client part $D$ with 40,000 examples and the holdout part $H$ with 10,000 examples. $D$ is partitioned into 10 clients with 1-class non-IID data. $H$ is used to create 10 random $G's$ with $\beta$ ranging from 2.5\% to 25\%. First, each entire $G$ is merged with the data of each client and 10 CNNs are trained by $FedAvg$ on the merged data from scratch over 300 communication rounds. The test accuracy is plotted against $\beta$ in Figure \ref{sol:figure7}. 
Second, we pick two specific $G's$, i.e., $G_{10\%}$ when $\beta = 10\%$ and $G_{20\%}$ when $\beta = 20\%$. For each $G$, (a) a warm-up CNN model is trained on $G$ to a test accuracy of \textasciitilde{60\%} (b) only a random $\alpha$ portion is merged with the data of each client and the warm-up model is trained on the merged data. The test accuracy is plotted against $\alpha$ in Figure \ref{sol:figure7}. The same training parameters for Section \ref{theory} are used.

As shown in Figure \ref{sol:figure7}(a), the test accuracy increases up to 78.72\% as $\beta$ increases. Even with a lower $\beta$ = 10\%, we can still achieve a test accuracy of 74.12\% for the extreme 1-class non-IID data, compared to 44\% without the {data-sharing} strategy. 
Moreover, it turns out that it is not necessary to distribute the entire $G$ to the clients to achieve a similar accuracy. Instead, only a random portion of $G$ needs to be distributed to each client. As shown in Figure \ref{sol:figure7}(b), the test accuracy of the warm-up model slowly increases with $\alpha$, up to 77.08\% for $G_{20\%}$ and 73.12\% for $G_{10\%}$. In particular, after an initially quick rise, the test accuracy changes less than 1\% for both $G_{20\%}$ and $G_{10\%}$ when $\alpha$ changes from 50\% to 100\%. Therefore, we can further reduce the size of the data that is actually received by each client with a proper selection of $\alpha$. For instance, if we are willing to centralize 10\% of the total clients' data in the cloud and randomly distribute 50\% of the globally shared data to the clients as well as a warm-up model at the initialization stage of federated learning, the test accuracy of $FedAvg$ can be increased by 30\% for the extreme 1-class non-IID data while only 5\% of the total data is actually received by each client. 

In summary, the {data-sharing} strategy offers a solution for federated learning with non-IID data. The size of the {globally shared dataset} and the random distributed fraction ($\alpha$) can be tuned depending on the problems and applications. The strategy only needs to be performed once when federated learning is initialized, so the communication cost is not a major concern. The globally shared data is a separate dataset from the clients' data so it is not privacy sensitive. 


\section{Conclusion}
\label{conclusion}
Federated learning will play a key role in distributed machine 
learning where data privacy is of 
paramount importance. Unfortunately, the quality of model 
training degrades if each of the edge devices sees
a unique distribution of data. In this work, we first show that 
the accuracy of federated learning reduces significantly,
by up to \textasciitilde55\% for neural networks trained on highly skewed non-IID data. We further
show that this accuracy reduction can be explained by the weight divergence, which can be quantified
by the earth movers distance (EMD) between the distribution over classes on each device and the
population distribution. As a solution, we propose a strategy to improve training on non-IID 
data by creating a 
small subset of data which is globally shared between all the edge devices. Experiments show 
that accuracy can be increased
by \textasciitilde30\% for the CIFAR-10 dataset with only 5\% globally shared data. 
There are still quite a few challenges left to make federated
learning mainstream, but improving model training on non-IID 
data is key to make progress in this area.

\newpage

{
\small
\bibliographystyle{ieeetr}
\bibliography{reference}
\bibliographystyle{plain}
}

\newpage

\appendix

\section{Appendix}
\label{apd}

\subsection{Test accuracy over communication rounds for a smaller batch size}
\label{apd:figure8}
\begin{figure}[ht]
	\centering
	\subfloat{\includegraphics[width=1.0\linewidth]{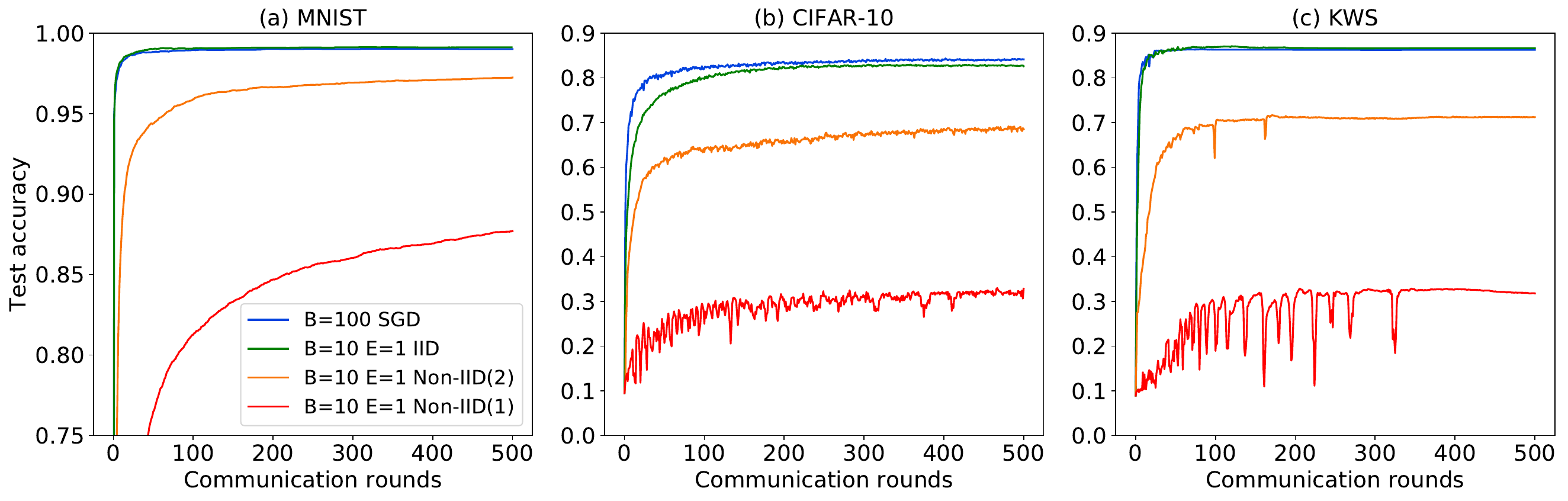}}
	\caption{Test accuracy over communication rounds of $FedAvg$ compared to SGD with IID and non-IID data of (a) MNIST (b) CIFAR-10 and (c) KWS datasets. Non-IID(2) represents the 2-class non-IID and non-IID(1) represents the 1-class non-IID. } 
\end{figure}

\subsection{Test accuracy of SGD and $FedAvg$ with IID or non-IID data}
\label{apd:table3}
\begin{table}[ht]
	\centering
	\caption{The test accuracy of SGD and $FedAvg$ with IID or non-IID data.}
	\begin{tabular}{|c c c c c c|}
		\hline
		Training & B & E & MNIST (\%)& CIFAR-10 (\%) & KWS (\%)\\
		\hline
		SGD & large & NA & 98.69 & 81.51 & 84.46 \\
		\hline
		FedAvg IID & large & 1 & 98.69 & 80.83 & 84.82 \\
		\hline
		FedAvg non-IID(2) & large & 1 & 96.29 & 67.00 & 72.30 \\
		\hline
		FedAvg non-IID(1) & large & 1 & 92.17 & 43.85 & 40.82 \\
		\hline 
		FedAvg non-IID(1) & large & 5 & 91.92 & 44.40 & 40.84 \\
		\hline
		Pre-trained non-IID(1) & large & 1 & 96.19 & 61.72 & 63.58 \\
		\hline
		SGD & small & NA & 99.01 & 84.14 & 86.28 \\
		\hline
		FedAvg IID & small & 1 & 99.12 & 82.62 & 86.64 \\
		\hline
		FedAvg non-IID(2) & small & 1 & 97.24 & 68.53 & 71.21\\
		\hline
		FedAvg non-IID(1) & small & 1 & 87.70 & 32.83 & 31.78\\
		\hline 
	\end{tabular}
\end{table}

\subsection{Proof of Proposition~\ref{prop:weight_div}}
\begin{proof}
	\label{apd:deriv}
Based on the definition of $\w^{(f)}_{mT}$ and $\w^{(c)}_{mT}$, we have

\begin{align*}
	  	 & || \w^{(f)}_{mT} - \w^{(c)}_{mT} || \\
	= 	 & || \sum_{k=1}^{K} \frac{n^{(k)}}{\sum_{k=1}^{K} n^{(k)}} \w^{(k)}_{mT} - \w^{(c)}_{mT} || \\
	= 	 & || \sum_{k=1}^{K} \frac{n^{(k)}}{\sum_{k=1}^{K} n^{(k)}} (\w^{(k)}_{mT-1} - \eta \sum_{i=1}^{C} p^{(k)}(y=i) \nabla_{\w} \E_{\x|y=i}[\log f_i(\x, \w^{(k)}_{mT-1})] \\
	  	 & \qquad - \w^{(c)}_{mT-1} + \eta \sum_{i=1}^{C} p(y=i) \nabla_{\w} \E_{\x|y=i}[\log f_i(\x, \w^{(c)}_{mT-1})]) \\
	\overset{1}{\leq} & || \sum_{k=1}^{K} \frac{n^{(k)}}{\sum_{k=1}^{K} n^{(k)}} \w^{(k)}_{mT-1} - \w^{(c)}_{mT-1} || \\
		 & \qquad + \eta || \sum_{k=1}^{K} \frac{n^{(k)}}{\sum_{k=1}^{K} n^{(k)}} \sum_{i=1}^{C} p^{(k)} (y=i)(\nabla_{\w} \E_{\x|y=i}[\log f_i(\x, \w^{(k)}_{mT-1})] - \nabla_{\w} \E_{\x|y=i}[\log f_i(\x, \w^{(c)}_{mT-1})]) || \\
	\overset{2}{\leq} & \sum_{k=1}^{K} \frac{n^{(k)}}{\sum_{k=1}^{K} n^{(k)}} (1 + \eta \sum_{i=1}^{C} p^{(k)}(y=i) \lambda_{\x|y=i}) || \w^{(k)}_{mT-1} - \w^{(c)}_{mT-1} ||.
\end{align*}

Here, inequality $1$ holds because for each class $i \in [C]$, $p(y=i) = \sum_{k=1}^{K} \frac{n^{(k)}}{\sum_{k=1}^{K} n^{(k)}} p^{(k)}(y=i)$, i.e., the data distribution over all the clients is the same as the distribution over the whole population. Inequality $2$ holds because we assume $\nabla_{\w} \E_{\x|y=i} [\log f_i(\x, \w)]$ is $\lambda_{\x|y=i}$-Lipschitz.

In terms of $|| \w^{(k)}_{mT-1} - \w^{(c)}_{mT-1} ||$ for client $k \in [K]$, we have

\begin{align}
\label{eq:local_weight_div}
	  & || \w^{(k)}_{mT-1} - \w^{(c)}_{mT-1} || \notag\\
	= & || \w^{(k)}_{mT-2} - \eta \sum_{i=1}^{C} p^{(k)}(y=i) \nabla_{\w} \E_{\x|y=i} [\log f_i(\x, \w^{(k)}_{mT-2})]  	  \notag\\
	  & \qquad - \w^{(c)}_{mT-2} + \eta \sum_{i=1}^{C} p(y=i) \nabla_{\w} \E_{\x|y=i} [\log f_i(\x, \w^{(k)}_{mT-2})] ||  \notag\\
	\leq & || \w^{(k)}_{mT-2} - \w^{(c)}_{mT-2} || + \eta || \sum_{i=1}^{C} p^{(k)}(y=i) \nabla_{\w} \E_{\x|y=i} [\log f_i(\x, \w^{(k)}_{mT-2})]  \notag\\
	  & \qquad - \sum_{i=1}^{C} p(y=i) \nabla_{\w} \E_{\x|y=i} [\log f_i(\x, \w^{(c)}_{mT-2})] ||  \notag\\
	\overset{3}{\leq} & || \w^{(k)}_{mT-2} - \w^{(c)}_{mT-2} || + \eta || \sum_{i=1}^{C} p^{(k)}(y=i) (\nabla_{\w} \E_{\x|y=i} [\log f_i(\x, \w^{(k)}_{mT-2})] - \nabla_{\w} \E_{\x|y=i} [\log f_i(\x, \w^{(c)}_{mT-2})] ) ||  \notag\\
	  & \qquad + \eta || \sum_{i=1}^{C} (p^{(k)}(y=i) - p(y=i)) \nabla_{\w} \E_{\x|y=i} [\log f_i(\x, \w^{(c)}_{mT-2})] || \notag\\
	\overset{4}{\leq} & (1 + \eta \sum_{i=1}^{C} p^{(k)}(y=i) L_{\x|y=i}) || \w^{(k)}_{mT-2} - \w^{(c)}_{mT-2} || + \eta g_{max}(\w^{(c)}_{mT-2}) \sum_{i=1}^{C} || p^{(k)}(y=i) - p(y=i) ||.
\end{align}

Here, inequality $3$ holds because

\begin{align*}
	& || \sum_{i=1}^{C} p^{(k)}(y=i) \nabla_{\w} \E_{\x|y=i} [\log f_i(\x, \w^{(k)}_{mT-2})] - \sum_{i=1}^{C} p(y=i) \nabla_{\w} \E_{\x|y=i} [\log f_i(\x, \w^{(c)}_{mT-2})] || \\
	= & || \sum_{i=1}^{C} p^{(k)}(y=i) \nabla_{\w} \E_{\x|y=i} [\log f_i(\x, \w^{(k)}_{mT-2})] - \sum_{i=1}^{C} p^{(k)}(y=i) \nabla_{\w} \E_{\x|y=i} [\log f_i(\x, \w^{(c)}_{mT-2})] + \\
	& \qquad \sum_{i=1}^{C} p^{(k)}(y=i) \nabla_{\w} \E_{\x|y=i} [\log f_i(\x, \w^{(c)}_{mT-2})] - \sum_{i=1}^{C} p(y=i) \nabla_{\w} \E_{\x|y=i} [\log f_i(\x, \w^{(c)}_{mT-2})] || \\
	\leq & || \sum_{i=1}^{C} p^{(k)}(y=i) (\nabla_{\w} \E_{\x|y=i} [\log f_i(\x, \w^{(k)}_{mT-2})] - \nabla_{\w} \E_{\x|y=i} [\log f_i(\x, \w^{(c)}_{mT-2})] ) || \\
	& \qquad + || \sum_{i=1}^{C} (p^{(k)}(y=i) - p(y=i)) \nabla_{\w} \E_{\x|y=i} [\log f_i(\x, \w^{(c)}_{mT-2})] ||.
\end{align*}

Inequality $4$ holds because $g_{max}(\w^{(c)}_{mT-2}) = \max_{i=1}^{C} || \nabla_{\w} \E_{\x|y=i} \log f_i(\x, \w^{(c)}_{mT-2}) ||$.

Based on Eq.~\eqref{eq:local_weight_div}, let $a^{(k)} = 1 + \eta \sum_{i=1}^{C} p^{(k)}(y=i) \lambda_{\x|y=i}$, by induction, we have

\begin{align*}
	& || \w^{(k)}_{mT-1} - \w^{(c)}_{mT-1} || \\
	\leq & a^{(k)} || \w^{(k)}_{mT-2} - \w^{(c)}_{mT-2} || + \eta g_{max}(\w^{(c)}_{mT-2}) \sum_{i=1}^{C} || p^{(k)}(y=i) - p(y=i) || \\
	\leq & (a^{(k)})^{2}   || \w^{(k)}_{mT-3} - \w^{(c)}_{mT-3} || + \eta \sum_{i=1}^{C} || p^{(k)}(y=i) - p(y=i) || (g_{max}(\w^{(c)}_{mT-2}) + a^{(k)} g_{max}(\w^{(c)}_{mT-3})) \\
	\leq & (a^{(k)})^{T-1} || \w^{(k)}_{(m-1)T} - \w^{(c)}_{(m-1)T} || + \eta \sum_{i=1}^{C} || p^{(k)}(y=i) - p(y=i) || (\sum_{j=0}^{T-2} (a^{(k)})^{j} g_{max}(\w^{(c)}_{mT-2-j}) ) \\
	= 	 & (a^{(k)})^{T-1} || \w^{(f)}_{(m-1)T} - \w^{(c)}_{(m-1)T} || + \eta \sum_{i=1}^{C} || p^{(k)}(y=i) - p(y=i) || (\sum_{j=0}^{T-2} (a^{(k)})^{j} g_{max}(\w^{(c)}_{mT-2-j}) ) \\
\end{align*}

Therefore,

\begin{align*}
	& || \w^{(f)}_{mT} - \w^{(c)}_{mT} || \\
	\leq & \sum_{k=1}^{K} \frac{n^{(k)}}{\sum_{k=1}^{K} n^{(k)}} ((a^{(k)})^{T} || \w^{(f)}_{(m-1)T} - \w^{(c)}_{(m-1)T} || \\
		 & \qquad + \eta \sum_{i=1}^{C} || p^{(k)}(y=i) - p(y=i) || (\sum_{j=0}^{T-1} (a^{(k)})^{j} g_{max}(\w^{(c)}_{mT-1-k}) )).
\end{align*}

Hence proved.

\end{proof}

\end{document}